\documentclass[letterpaper, 10 pt, conference]{ieeeconf} 
\IEEEoverridecommandlockouts
\usepackage{cite}
\usepackage{comment}
\usepackage{color}
\usepackage{soul}
\usepackage{amsmath}
\usepackage{amssymb}
\usepackage{amsfonts}

\usepackage[font=footnotesize]{subcaption}
\usepackage[font=footnotesize]{caption}

\usepackage{graphicx}
\usepackage{float}
\usepackage{amsmath}
\usepackage{multirow}
\usepackage[ruled,linesnumbered]{algorithm2e}
\usepackage{enumerate}
\usepackage{amssymb}
\usepackage{comment}
\usepackage{caption}

\newcommand{\argmin}{\mathop{\mathrm{arg~min}}\limits}

\SetAlCapNameFnt{\small} 
\SetAlCapFnt{\small}

\title{\LARGE \bf
Feasibility-aware Imitation Learning from Observations \\ through a Hand-mounted Demonstration Interface
}

\author{
    Kei Takahashi$^{1}$, 
    Hikaru Sasaki$^{1}$ and Takamitsu Matsubara$^{1}$
    \thanks{
        *This work was supported by JSPS KAKENHI Grant Numbers JP24KJ1702 and JP24K03018.
    }
    \thanks{
        $^{1}$K. Takahashi, H. Sasaki, and T. Matsubara are with the Division of Information Science, Graduate School of Information Science, Nara Institute of Science and Technology (NAIST), Nara, Japan.{\tt\small}
    }%
}

\begin{document}
\maketitle
\thispagestyle{empty}
\pagestyle{empty}

\begin{abstract}
Imitation learning through a demonstration interface is expected to learn policies for robot automation from intuitive human demonstrations.
However, due to the differences in human and robot movement characteristics, a human expert might unintentionally demonstrate an action that the robot cannot execute.
We propose feasibility-aware behavior cloning from observation (FABCO).
In the FABCO framework, the feasibility of each demonstration is assessed using the robot's pre-trained forward and inverse dynamics models.
This feasibility information is provided as visual feedback to the demonstrators, encouraging them to refine their demonstrations.
During policy learning, estimated feasibility serves as a weight for the demonstration data, improving both the data efficiency and the robustness of the learned policy.
We experimentally validated FABCO's effectiveness by applying it to a pipette insertion task involving a pipette and a vial.
Four participants assessed the impact of the feasibility feedback and the weighted policy learning in FABCO. Additionally, we used the NASA Task Load Index (NASA-TLX) to evaluate the workload induced by demonstrations with visual feedback.
\end{abstract}

\section{Introduction}
Imitation learning \cite{osa2018} is a method for learning policies that assign automation using human demonstrations rather than manually designing them through environmental models.
Recently, approaches have been explored that learn policies for robotic automation only from environmental observations collected during direct demonstrations by a human expert.
To simplify policy learning in such imitation learning, demonstration interfaces that mimic the structure of a robot's hand have been proposed \cite{chi2024universal, shafiullah2023bringing, hamaya2020, young2021}.

Imitation learning with a demonstration interface cannot directly learn a policy since the training data do not include action information.
To address this issue, we focus on imitation learning from observation (ILfO), an imitation learning method that assumes that the only available option is the observation of demonstrations and incorporates a mechanism to complement the action information.
ILfO assumes that a policy is learned using the robot's motion data.
An ILfO method, behavior cloning from observation (BCO), learns an inverse dynamics model (IDM) using the robot's motion data and a policy by predicting the action based on the demonstration data by IDM\cite{torabi2018}.
Reinforcement learning-based methods that use demonstration data as rewards have also been proposed\cite{liu2018imitation, rafailov2021, alejandro2023}.

Imitation learning with a demonstration interface allows human experts to intuitively demonstrate actions without prior robotics knowledge.
However, due to differences in human and robot movement characteristics, a human expert may demonstrate an infeasible action for a robot.
A policy learned using such demonstrations cannot achieve the task, and the error between the behaviors of the policy and the demonstration is compounded by a covariate shift \cite{ross2010}.

\begin{figure}[t]
    \centering
    \includegraphics[width=0.95\hsize]{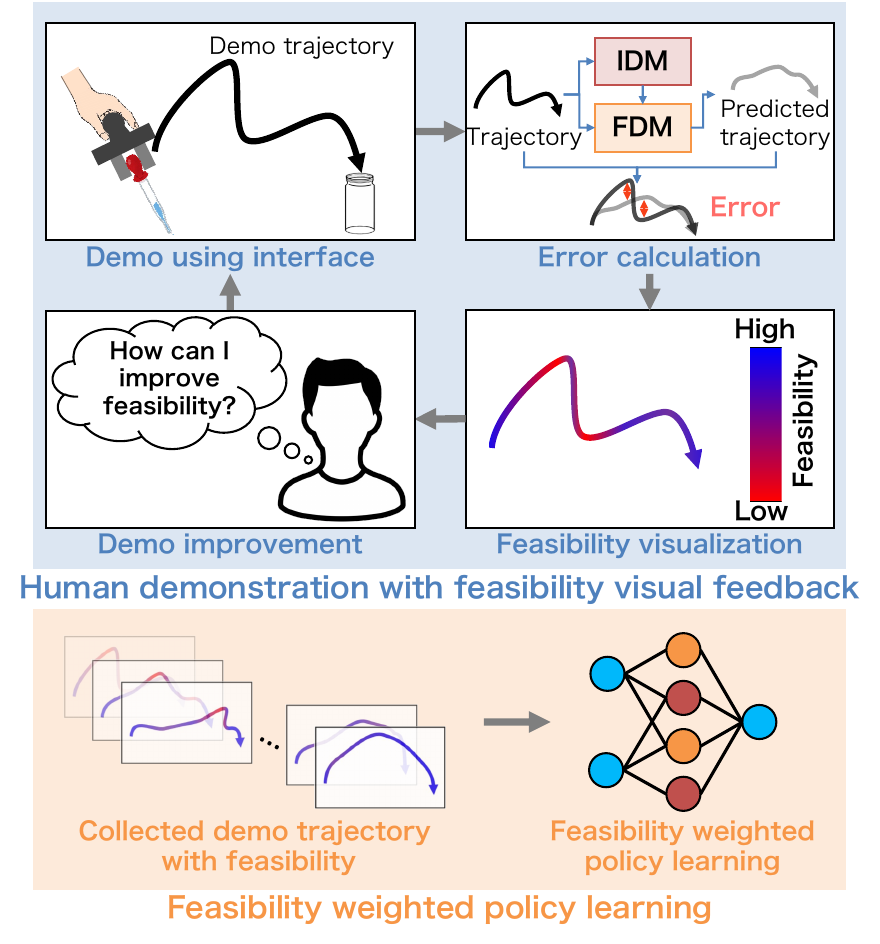}
    \caption{Overview of FABCO: Demonstrator performs demonstrations using interface and receives visual feedback on feasibility of robot's execution. This encourages demonstrator to provide demonstrations with higher feasibility. Collected demonstration trajectory is weighted based on feasibility to learn a robust policy.}
    \label{fig:proposed:overview}
\end{figure}

We believe that the above problems can be alleviated by introducing a robot's feasibility into imitation learning with a demonstration interface and using it to provide feedback to the human expert and the policy learning.
Since feasibility can be calculated using the IDM and the forward dynamics model (FDM), it can be introduced to BCO by simultaneously training the FDM and the IDM.

In this study, we propose a novel ILfO method called feasibility-aware behavior cloning from observation (FABCO) that has a demonstration interface.
In the FABCO framework, the feasibility of the demonstration is evaluated using the robot's pre-trained FDM and IDM.
This feasibility is presented as visual feedback to the demonstrator, encouraging clarification of the demonstration data. During policy learning, the estimated feasibility is used as a weight to the demonstration data, enhancing both the data efficiency and the robustness of the learned policy.
An overview of FABCO is shown in Fig. \ref{fig:proposed:overview}.
We verify the effectiveness of the proposed method by applying it to a pipette insertion task using a pipette and a vial.
In this experiment, four participants verify the effectiveness of feasibility visual feedback and feasibility-weighted policy learning on FABCO.
We also used NASA-TLX \cite{HART1988139} to investigate the workload induced by demonstrations using visual feedback.

\section{Related Work}
\subsection{Imitation Learning using Robot Demonstrations}
Imitation learning methods acquire a control policy using demonstration data, which
are collected through a robot operation.
Wong et al. proposed a system that enables users to control a mobile robot by a smartphone while viewing images from a camera mounted on the robot. This situation allows for the teaching of both mobility operations and robotic arm manipulation for imitating kitchen tasks \cite{wong2022error}.
Cuan et al. demonstrated that adding real-time haptic feedback to a teleoperation device improves the efficiency and quality of the data collection and imitated a door-opening task\cite{cuan2024leveraginghapticfeedbackimprove}.
However, in these imitation learning methods, the learned policy is known to be vulnerable to modeling errors and noise, which can cause the robot's behavior based on the learned policy to become unstable.
This problem is called a covariate shift.
Interactive approaches have been proposed to address this issue.

Interactive imitation learning alternates between demonstration and policy learning and acquires a stable policy by utilizing a learned policy for demonstration.
Two representative methods are DAgger and DART.
In the former approach, the robot moves according to a learned policy, and the operator takes action and collects data only when learning is insufficient \cite{Ross2010ARO, michael2016, michael2019, kunal2019, mandlekar2020humanintheloopimitationlearningusing}.
The DART approach injects noise into human operations and prompts human operators to take action to deal with it \cite{laskey2017, tahara2022, tahara2023, oh2023, Oh_2024}.

These imitation learning methods assume that demonstration data are collected by directly operating a robot.
However, operating it requires specialized skills.
To enable robots to automate a broader range of tasks, methods are desirable that allow them to learn from demonstrations without requiring such specialized skills.

\subsection{Imitation Learning using Human Hand Demonstrations}
Imitation learning using human hand demonstrations allows human experts to perform intuitive motion demonstrations.
However, the corresponding robot states and actions must be estimated from human demonstration data since the states and actions of humans and robots are different.
Some methods use trial and error to estimate actions that imitate the estimated robot state sequences \cite{liu2018imitation, smith2019avid, qin2021from}.
Smith et al. used a CycleGAN model to replace humans with robots in demonstration videos and learned corresponding robot actions through model-based reinforcement learning \cite{smith2019avid}.
While those studies provide intuitive demonstration, this approach struggles with the physical differences between humans and robots.

Recent research has focused on unifying human and robot end-effectors to simplify imitation \cite{chi2024universal, shafiullah2023bringing, hamaya2020, young2021}. 
A previous work developed a demonstration interface that unified human and robot end-effectors, enabling through behavior cloning the imitation of tasks like grasping and pushing \cite{young2021}. 
Chi et al. also developed a similar demonstration interface that facilitated the learning of generalizable policies \cite{chi2024universal}. 

In contrast to those studies, we unify end-effectors using a hand-mounted demonstration interface and provide feasibility visual feedback for demonstrators and robot-executable demonstrations.

\subsection{Feasibility in Imitation Learning}
In the imitation learning framework, the differences between the demonstrator and robot dynamics are addressed by estimating feasibility, which is utilized for policy learning \cite{cao2021corl, cao2021ral, Betz_2021}.
Since these methods evaluate feasibility after demonstrations, they may collect many demonstrations with low feasibility.
Our method introduces feasibility into ILfO and proposes a method that incorporates it into feedback to the demonstrator and policy learning.

\section{Behavior Cloning from Observation}
BCO is an imitation learning method inspired by learning through human imitation. 
It acquires a policy that imitates the demonstrator's behavior using only the state sequences observed by the demonstrator during task execution \cite{torabi2018}. 
This method compensates for the lack of action information in the demonstration data by learning the agent's IDM.
It predicts the agent's actions based on the state transitions in the demonstration data and learns the policy accordingly.

\subsection{Problem Statement}
BCO assumes that the environment follows a Markov decision process (MDP). At time $t=1$, initial state $\mathbf s_1$ follows initial state distribution $p(\mathbf s_1)$, and at time $t$, given state $\mathbf s_t$ and action $\mathbf a_t$, subsequent state $\mathbf s_{t+1}$ follows state transition probability $p(\mathbf s_{t+1}\mid \mathbf s_t, \mathbf a_t)$. 
The policies of the demonstrator and the agent are denoted as $\pi^e(\mathbf a_t\mid \mathbf s_t)$ and $\pi^r_\phi(\mathbf a_t\mid \mathbf s_t)$. 
While the demonstrator provides demonstrations based on $\pi^e(\mathbf a_t\mid \mathbf s_t)$, during policy learning for the agent, only state sequence $\mathbf S^e \! = \! \{\mathbf s_t^e\}_{t=1}^T$ from the demonstrations is available.

\subsection{IDM Learning and Prediction of Demonstrator's Actions}
BCO utilizes IDM to predict actions from demonstration state sequence $\mathbf S^e$.
The IDM is trained using state sequence $\mathbf S^r \! = \! \{\mathbf s_t^r\}_{t=1}^T$ and action sequence $\mathbf A^r \! = \! \{\mathbf a_t^r\}_{t=1}^T$ collected when the agent follows arbitrary policy $\pi^\mathrm{rand}$ in the environment.
IDM $f^\mathrm{IDM}_\theta$ predicts action $\mathbf a_t$ using states $\mathbf s_t$ and $\mathbf s_{t+1}$, and assuming that the agent's state sequence $\mathbf S^r$ and action sequence $\mathbf A^r$ follow $p(\mathbf S^r, \mathbf A^r\mid \pi^\mathrm{rand})$, IDM parameters $\theta$ are optimized:
\begin{align}
    \theta^* = \argmin_\theta \mathbb E_{p(\mathbf  S^r, \mathbf A^r\mid \pi^\mathrm{rand})}\left[ \sum_{t=1}^{T-1} \left| \mathbf{a}_t^r - f^\mathrm{IDM}_\theta(\mathbf{s}_t^r, \mathbf{s}_{t+1}^r) \right|\right].
\end{align}
Using optimized parameters $\theta^*$, action $\tilde{\mathbf a}^e_t=f^\mathrm{IDM}_{\theta^*}(\mathbf s^e_t,\mathbf s^e_{t+1})$ is predicted for demonstration states $\mathbf s^e_t$ and $\mathbf s^e_{t+1}$.

\subsection{Policy Learning}
Parameters $\phi$ of agent's policy $\pi_\phi^r$ are learned using demonstration state sequence $\mathbf S^e$ and IDM. 
When demonstration state sequence $\mathbf S^e$ follows $p(\mathbf S^e\mid \pi^\mathrm{e})$, the parameters are learned:
\begin{align}
    \phi^* = \argmin_\phi \mathbb E_{p(\mathbf S^e\mid \pi^\mathrm{e})}\left[ \sum_{t=1}^{T-1} \left| f^\mathrm{IDM}_{\theta^*}(\mathbf s^e_t,\mathbf s^e_{t+1}) - \pi_\phi^r (\mathbf s^e_t) \right| \right].
\end{align}

\section{Feasibility-aware Behavior Cloning from Observation}
In this section, we propose FABCO, a novel ILfO method with a hand-mounted demonstration interface that combines visual feasibility feedback for the demonstrators and feasibility-weighted policy learning for the imitation learning (Fig. \ref{fig:proposed:overview}). 
FABCO is based on the assumption that motion data can be collected from a robot's random actions. 
It evaluates the feasibility of demonstrations by learning FDM and IDM with the robot's motion data.

\subsection{Hand-mounted Demonstration Interface}
\begin{figure}[t]
    \centering
    \begin{minipage}[b]{0.45\linewidth}
        \centering
        \includegraphics[width=\hsize]{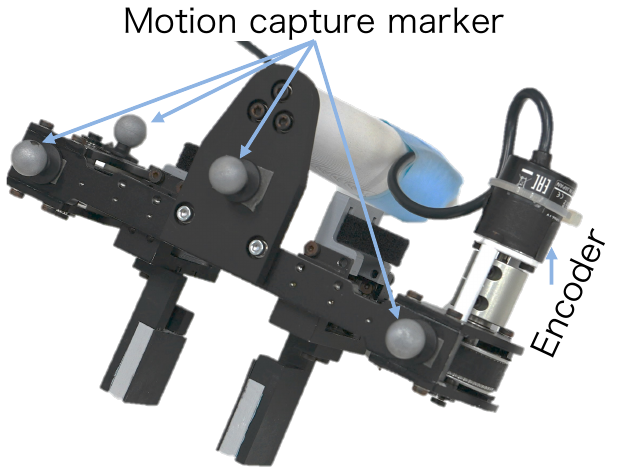}
        \subcaption{}
    \end{minipage}
    \begin{minipage}[b]{0.45\linewidth}
        \centering
        \includegraphics[width=\hsize]{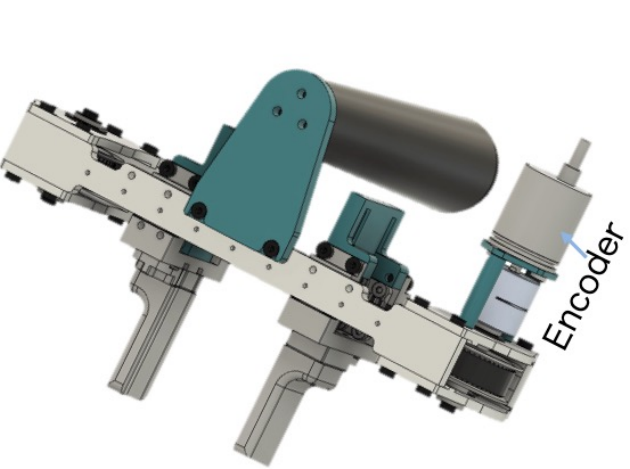}
        \subcaption{}
    \end{minipage}
    \caption{Hand-mounted demonstration interface: (a) its overview and (b) its CAD design.}
    \label{fig:experiment:common_device}
\end{figure}

To collect demonstration data with a shared hand structure between a robot and a human, we developed a hand-mounted demonstration interface that shares the two-finger opposing structure of the robot hand with the human hand (Fig. \ref{fig:experiment:common_device}). 
This hand-mounted demonstration interface, which was developed with reference to Hamaya et al. \cite{hamaya2020}, allows the demonstrator to grasp the handle and open/close the fingers. 
The distance between the fingers can be calculated from an encoder attached to the hand-mounted demonstration interface, and the interface's pose is observed using a motion capture system.

\subsection{Problem Statement}
Demonstration data $\mathbf S^e \!= \! \{\mathbf s^e_t\}_{t=1}^T$ from the task execution consists of state $\mathbf s^e_t$, which is composed of hand-mounted demonstration interface pose information $\mathbf p^e_t$ and environmental observation $\mathbf o^e_t$. This method focuses on capturing the robot's dynamics and assumes that the robot's actions can be estimated from the sequence of its poses $\mathbf p^r_t$ to acquire a generalizable model for the task. 
For IDM and FDM learning, the hand pose information from the demonstrations is used to predict the actions and the next hand pose. 
The training data for IDM and FDM consist of the robot's pose series data $\mathbf P^r \!= \!\{\mathbf p^r_t\}_{t=1}^T$ and action series data $\mathbf A^r=\{\mathbf a^r_t\}_{t=1}^T$, collected as the robot follows randomly generated motion trajectories within the workspace using tracking policy $\pi^\mathrm{track}$. For policy learning, the actions are predicted from the pose transitions included in demonstration data $\mathbf S^e$, and the policy is learned based on these predictions.

\subsection{Learning of Inverse and Forward Dynamics Models}
IDM and FDM are trained using data collected when the robot moves randomly within the workspace. 
The training data for IDM and FDM are collected by generating random motion trajectories within the robot's operational environment. 
The robot follows these trajectories using a tracking policy $\pi^\mathrm{track}$, collecting a series of robot pose data $\mathbf P^r$ and action data $\mathbf A^r$ during the trajectory tracking.
Here robot's pose series data $\mathbf P^r$ and action series data $\mathbf A^r$ are assumed to follow $p(\mathbf P^r,\mathbf A^r\mid \pi^\mathrm{track})$.

IDM is a model, $f_\theta^\mathrm{IDM}(\mathbf p_t, \mathbf p_{t+1})$, which predicts action $\mathbf a_t$ required to reproduce a change in the pose between $\mathbf p_t$ and $\mathbf p_{t+1}$. 
FDM is a model, $f_\psi^\mathrm{FDM}(\mathbf p_t, \mathbf a_t)$, which estimates pose $\mathbf p_{t+1}$ that results from pose $\mathbf p_t$ and action $\mathbf a_t$. 
Parameters $\theta$ and $\psi$ of IDM and FDM are optimized based on the following equations:
\begin{align}
    \theta_* &= \argmin_\theta \mathbb E_{p(\mathbf P^r,\mathbf A^r\mid \pi^\mathrm{track})}\left[ \sum_{t=1}^{T-1} \left| \mathbf a_t^r - f^\mathrm{IDM}_\theta(\mathbf p_t^r, \mathbf p_{t+1}^r)\right| \right], 
    \label{eq:IDM} \\
    \psi_* &= \argmin_\psi \mathbb E_{p(\mathbf P^r,\mathbf A^r\mid \pi^\mathrm{track})}\left[ \sum_{t=1}^{T-1} \left| \mathbf p_{t+1}^r - f^\mathrm{FDM}_\psi(\mathbf p_t^r, \mathbf a_t^r)\right| \right].
    \label{eq:FDM}
\end{align}
Using the optimized parameters, IDM and FDM predict the action that changes the robot's pose, $\tilde{\mathbf a}_t = f_{\theta^*}^\mathrm{IDM}(\mathbf p_t, \mathbf p_{t+1})$, and the robot's pose at the next time step, $\tilde{\mathbf p}_{t+1} = f_{\psi^*}^\mathrm{FDM}(\mathbf p_t, \mathbf a_t)$.

\subsection{Feasibility Calculation}
\begin{figure}[t]
    \centering
    \includegraphics[width=0.8\hsize]{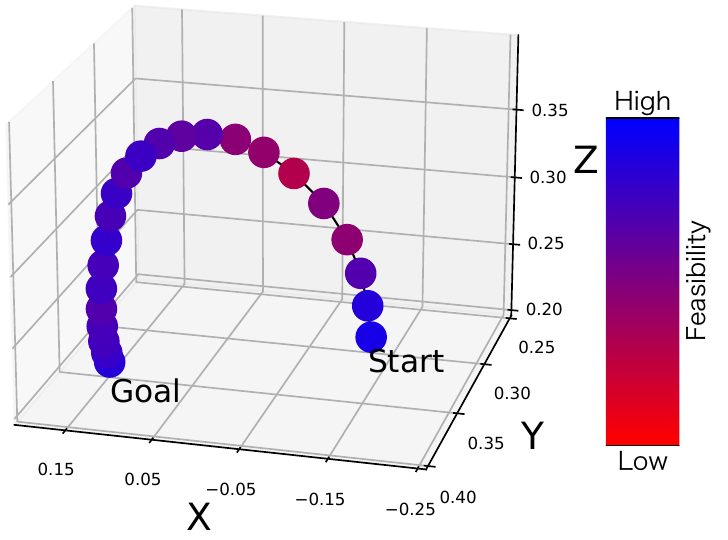}
    \caption{Visualization of feasibility feedback: Demonstration trajectory is color-coded based on feasibility and displayed with automatic rotation to encourage improvement in demonstrations.}
    \label{fig:visualization_of_feasibility}
\end{figure}

The feasibility of demonstrated hand poses $\mathbf P^e \! = \! \{\mathbf p^e_t\}_{t=1}^T$ is calculated and fed back to the demonstrator to encourage demonstrations that are easier for the robot to imitate. 
The feasibility is calculated using action prediction $\tilde{\mathbf a}_t^e = f^\mathrm{IDM}_{\theta^*}(\mathbf p_t^e, \mathbf p_{t+1}^e)$ from the IDM and the predicted robot pose at the next time step using FDM.
Feasibility $w_t$ at time step $t$ is calculated as follows:
\begin{align}
    w_t =\exp\left\{-\frac{|f^\mathrm{FDM}_{\theta^*}(\mathbf p_t^e, \tilde{\mathbf a}_t^e) - \mathbf p^e_{t+1}|}{2 \sigma^2_w}\right\}.
    \label{eq:feasibility}
\end{align}
Here $\sigma_w$ is a parameter that adjusts the scale of feasibility. Hand pose information $\mathbf P^e$ from the demonstrated motion is plotted on a graph, and feasibility $w_t$ is visualized by coloring the plot according to the values of $w_t$ (Fig. \ref{fig:visualization_of_feasibility}) to provide visual feedback to the demonstrator.

\subsection{Feasibility-weighted Policy Learning}
A policy is learned using demonstration data $\mathbf S^e$, a trained IDM, and feasibility $w_t$ of the demonstration data, where higher feasibility actions are weighted. 
Parameters $\phi$ of policy $\pi_\phi$ are learned using the following equation:
\begin{align}
    J_{\pi_\phi} &= \mathbb E_{p(\mathbf S^e\mid \pi^\mathrm{demo})}\left[\sum^{T-1}_{t=1}  w_t \left| f^\mathrm{IDM}_{\theta^*}(\mathbf p_t^e, \mathbf p_{t+1}^e) - \pi_\phi (\mathbf s^e_t) \right| \right],
    \label{eq:policy1} \\
    \phi^* &= \argmin_\phi J_{\pi_\phi}.
    \label{eq:policy2}
\end{align}
When using learned policy $\pi_{\phi^*}$, state $\mathbf s^r_t = \{\mathbf p^r_t, \mathbf o_t\}$, which includes robot's pose $\mathbf p^r_t$ and environmental information $\mathbf o_t$, is provided as input to the policy.
FABCO's learning process is shown in Algorithm \ref{alg:method}.

\begin{algorithm}[t]
    \footnotesize
    \SetAlgoLined
    \DontPrintSemicolon
    Initialize IDM $f_{\theta}^\mathrm{IDM}$, FDM $f_{\psi}^\mathrm{FDM}$ and, policy $\pi_{\phi}$\;
    \# Learning process of IDM and FDM \;
    Generate robot's trajectories $\mathcal D=\{\mathbf P^r_n, \mathbf A^r_n\}_{n=1}^N$ using trajectory tracking policy $\pi^\mathrm{track}$ \;    
    Learn IDM $f_{\theta}^\mathrm{IDM}$ using data $\mathcal D$ by Eq. \ref{eq:IDM}  \;
    Learn FDM $f_{\psi}^\mathrm{FDM}$ using data $\mathcal D$ by Eq. \ref{eq:FDM}  \;
    \# Human demonstration process\;
    \For{$m = 1$ \textbf{to} $M$}{
        Collect demonstration $\mathbf S^e_m = \{\mathbf P^e_m, \mathbf O^e_m\}$\;
        Predict action $\tilde{\mathbf A}^e_m$ for robot pose $\mathbf P^e_m$ using IDM\;
        Predict the transition of robot pose $\tilde{\mathbf P}^e_m$ for predicted action $\tilde{\mathbf A}^e_m$ and robot pose $\mathbf P^e_m$ using FDM\;
        Calculate feasibility $\mathbf w_m$ using $\mathbf P^e_m$ and $\tilde{\mathbf P}^e_m$ by Eq. \ref{eq:feasibility}\;
        Visual feedback of feasibility $\mathbf w_m$ to the expert\;
    }
    \# Feasibility-weighted policy learning process\;
    Learn $\pi_{\phi}$ using $\{\mathbf S^e_m$, $\tilde{\mathbf A}^e_m$, $\mathbf w_m\}_{m=1}^M$ by Eq. \ref{eq:policy2} \;
    \caption{Learning process of FABCO}
    \label{alg:method}
\end{algorithm}

\section{Experiment}
In this study, we apply our proposed method, FABCO, to a pipette insertion task.
We employ BCO as a comparison and conduct ablation studies with FABCO without visual feedback and without weighting.
Four participants demonstrated the task, and the proposed method's effectiveness was evaluated based on their demonstrations.
Additionally, we compared the workload on the demonstrators with/without visual feedback based on the NASA-TLX results collected after the demonstration tasks were completed.

\subsection{Pipette Insertion Task}
\begin{figure}[t]
    \centering
    \includegraphics[width=0.8\hsize]{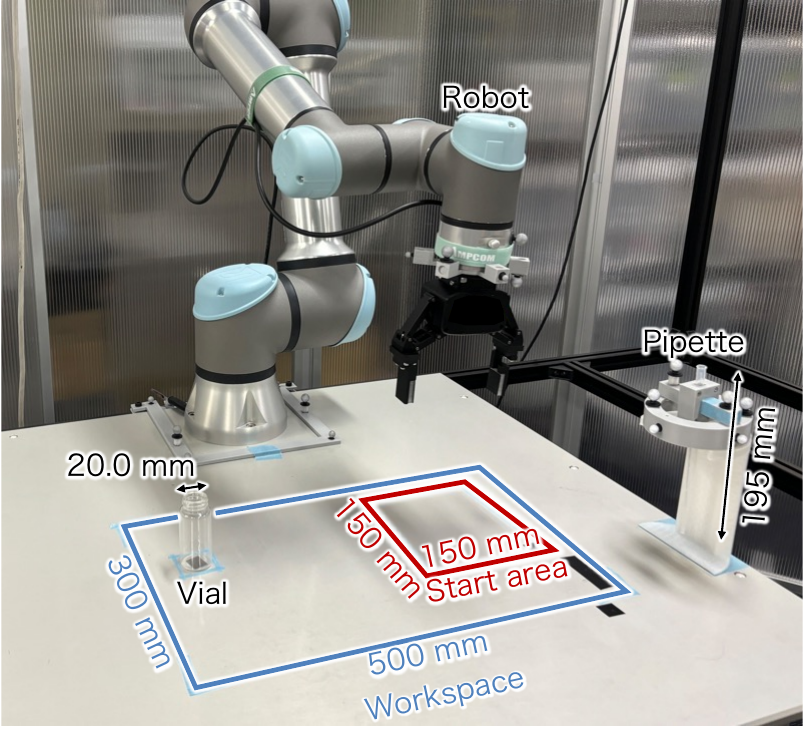}
    \caption{Pipette insertion task environment}
    \label{fig:experiment:environment}
\end{figure}

The experimental setup for the pipette insertion task is shown in Fig. \ref{fig:experiment:environment}. 
Its goal is for the robot to insert a pipette, which it is gripping, into a vial.
The pipette is 195 mm long, and the diameter of the vial's opening is 20.0 mm. 
The robot begins the task while holding the pipette in the start area, and its initial pose is randomized within a range of $\pm$ 15.0 degrees for each of the xyz axes in the Euler angles. 
The state of this task ($\mathbf s_t$) is represented by 12-dimensional data consisting of the 6-dimensional position and the orientation data of the robot (or the interface) and the 6-dimensional position and the orientation data of the pipette.
The robot's action ($\mathbf a_t$) is a 6-dimensional velocity command applied to the position and the orientation of the robot's end-effector.

In this experiment, we used UR5e from UNIVERSAL ROBOTS as the manipulator and a ROBOTIS HAND RH-P12-RN-UR from ROBOTIS as the robot hand. 
To measure the pose of the hand-mounted demonstration interface and the robot's hand, we employed six Flex13 cameras from OptiTrack. 
The workspace in this experiment, limited by the robot's range of motion, was 300 mm vertically, 500 mm horizontally, and 450 mm high. 
The angular range was set to ±30.0 degrees for rotation around the x-axis, $\pm$ 30.0 degrees for rotation around the y-axis, and $\pm$ 45.0 degrees for rotation around the z-axis.

\subsection{Learning Settings of IDM and FDM}
To train IDM and FDM, we generated random motion trajectories within the workspace shown in Fig. \ref{fig:experiment:environment} and collected training data. 
Five points were randomly sampled from the workspace, and the robot followed a trajectory that connected them.
Each trajectory consists of 50 position and orientation points;  the robot followed one trajectory to collect a series of 50 state-action data points.
The training data were collected by running the robot for approximately 2.5 hours, obtaining data from 2500 trajectories.

The IDM takes hand poses ${\mathbf p_{t-1}, \mathbf p_t, \mathbf p_{t+1}}$ as input and outputs action $\mathbf a_{t}$.
The FDM takes hand pose $\mathbf p_t$ and action $\mathbf a_t$ as input and outputs the hand pose at the next time step, $\mathbf p_{t+1}$. 
Both IDM and FDM are modeled using neural networks with five hidden layers, consisting of 256, 1024, 2048, 1024, and 256 nodes. 
ReLU is used as an activation function for the hidden layers; Sigmoid is used for the output layer. 
The optimizer is Adam, the loss function is L1 loss, and the batch size is set to 256. 
Out of the 2500 trajectories, 2000 were used as training data and 500 as validation data. 
The model was trained for 2000 epochs, and we selected the model with the highest prediction performance on the validation data at the end of each epoch.

\subsection{Human Demonstration}
\begin{figure}[!t]
    \centering
    \includegraphics[width=1\hsize]{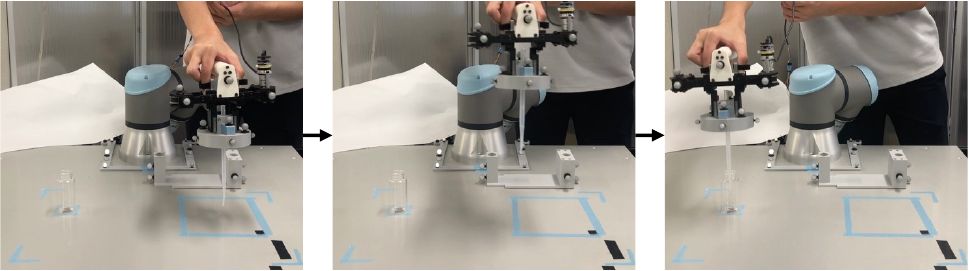}
    \caption{Pipette insertion motion in demonstration}
    \label{fig:experiment:expert_insert}
    \vspace{3mm}
    \centering
    \includegraphics[width=1\hsize]{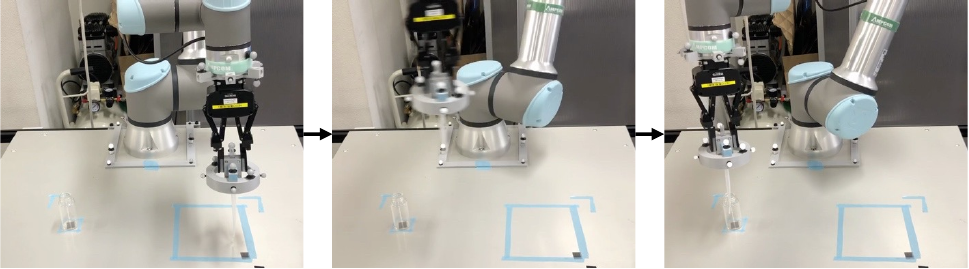}
    \caption{Pipette insertion motion by learned policy}
    \label{fig:experiment:robot_insert}
\end{figure}

We collected  pipette insertion demonstrations using the hand-mounted demonstration interface . 
The process of inserting the pipette into the vial is shown in Fig. \ref{fig:experiment:expert_insert}.
The demonstrators were informed in advance about the robot's operational range and speed and were instructed to demonstrate the task in such a way to complete it as quickly as possible within the robot's allowable range of motion and speed.
The initial states of the demonstrations were selected to ensure diversity within the experimental setup. 
In this experiment, we collected 50 trajectories each for demonstrations with FABCO visual feasibility feedback and demonstrations without visual feedback.
After the two types of demonstrations, the participants completed the NASA-TLX\cite{HART1988139} to evaluate their workload during the experiment.

\subsection{Policy Learning}
The policy model used for all the methods is neural networks
with two hidden layers, with 256 and 128 nodes in the first and second hidden layers.
ReLU was used as the activation function for both the hidden layers and the output layer. 
In this experiment, parameter $\sigma_w$, which adjusts the scale of feasibility, was set to $0.15$.
The optimizer is Adam; the loss function is L1 loss, with a batch size of 256. 
The model was trained for 2000 epochs, and the model was selected with the highest prediction performance on the validation data at the end of each epoch.

\subsection{Results}
Figure. \ref{fig:experiment:robot_insert} shows the robot performing a pipette insertion using the policy learned with FABCO, based on demonstrations provided by Subject 1.
It successfully imitated the pipette insertion task (Fig. \ref{fig:experiment:expert_insert}). 

Table \ref{table:experiment:success_rate} presents the success rates of the pipette insertion task, tested 30 times for the four methods (FABCO, FABCO w/o weighting, FABCO w/o FB, and BCO) that were trained using demonstrations from four participants. 
When comparing the success rates, the policy performance clearly improved with the visual feasibility feedback and weighting based on feasibility during the demonstrations. 
Notably, the visual feasibility feedback significantly improved the performance. 
Subject 2 tends to achieve a high task success rate regardless of the presence/absence of feedback.

\begin{table}[t]
    \centering
    \caption{Success rates for pipette insertion tasks learned from each subject's demonstration data: Task success rates are calculated using 30 executions for each method. Method with highest success rate for each subject is highlighted in bold.}
    \label{table:experiment:success_rate} 
    \begin{tabular}{|c||c|c|c|c|} \hline
        \begin{tabular}[c]{@{}c@{}}Human\\ participants\end{tabular} &
        FABCO &
        \begin{tabular}[c]{@{}c@{}}FABCO\\ w/o weighting\end{tabular} &
        \begin{tabular}[c]{@{}c@{}}FABCO\\ w/o FB\end{tabular} &
        BCO  \\ \hline\hline
        No. 1 & \textbf{93.3}\% & \textbf{93.3}\% &         56.7\%  & 53.3\% \\ \hline
        No. 2 &         96.7 \% & 96.7         \% & \textbf{100}\%  & 86.7\% \\ \hline
        No. 3 & \textbf{93.3}\% & 90.0         \% &         66.7\%  & 26.7\% \\ \hline
        No. 4 & \textbf{90.0}\% & 60.0         \% &         26.7\%  & 20.0\% \\ \hline
        \end{tabular}
\end{table}
\begin{figure}[!t]
    \centering
    \begin{minipage}[b]{1\linewidth}
        \centering
        \includegraphics[width=0.4\hsize]{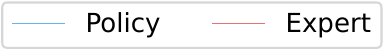}
    \end{minipage}
    \begin{minipage}[b]{0.48\linewidth}
        \centering
        \includegraphics[width=1\hsize]{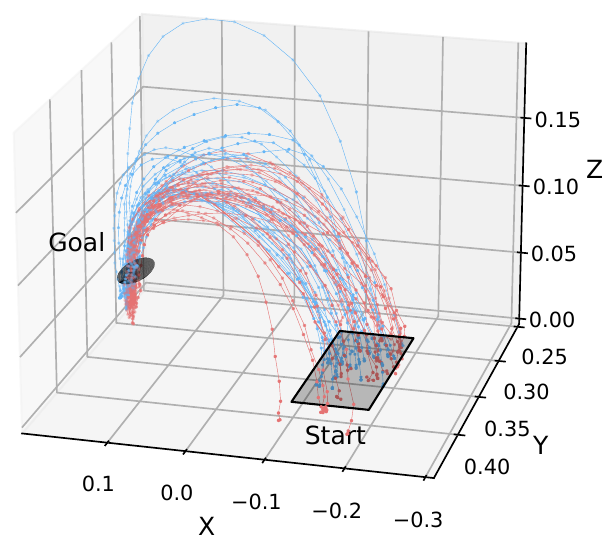}
        \vspace{-6mm}
        \subcaption{FABCO}
        \label{fig:experiment:trajectory_of_No.4_F&W}
    \end{minipage}
    \begin{minipage}[b]{0.48\linewidth}
        \centering
        \includegraphics[width=1\hsize]{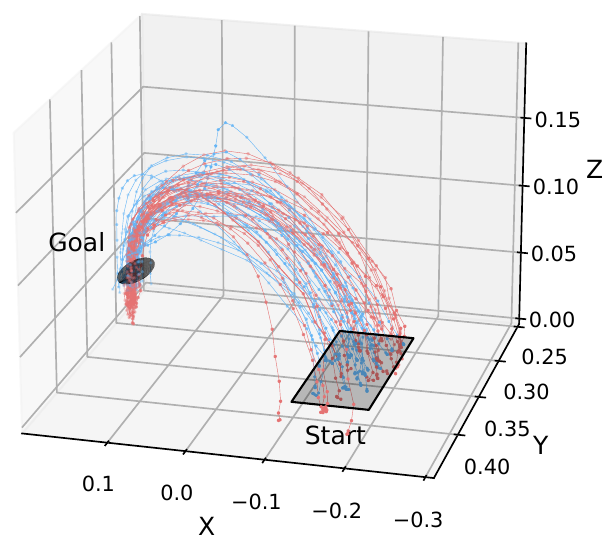}
        \vspace{-6mm}
        \subcaption{FABCO w/o weighting}
        \label{fig:experiment:trajectory_of_No.4_NoF&W}
    \end{minipage}
    \begin{minipage}[b]{0.48\linewidth}
        \centering
        \includegraphics[width=1\hsize]{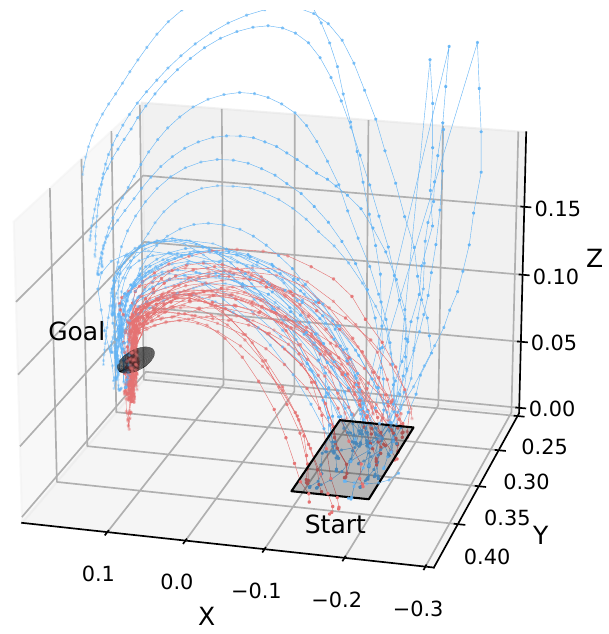}
        \vspace{-6mm}
        \subcaption{FABCO w/o FB}
        \label{fig:experiment:trajectory_of_No.4_F&NoW}
    \end{minipage}
    \begin{minipage}[b]{0.48\linewidth}
        \centering
        \includegraphics[width=1\hsize]{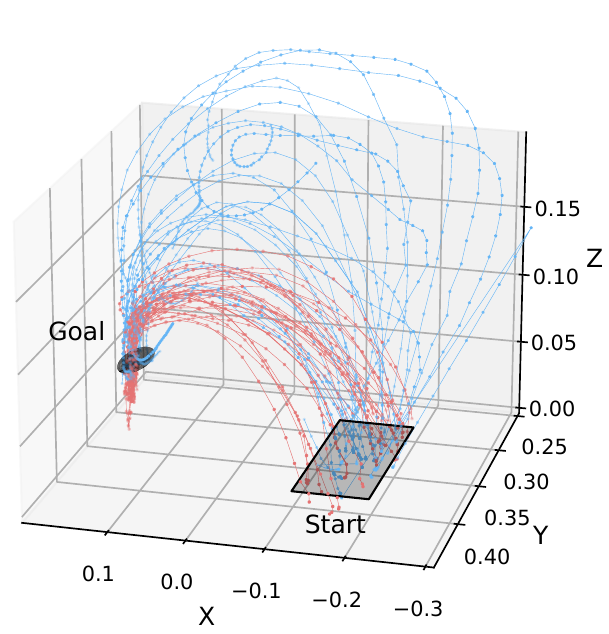}
        \vspace{-6mm}
        \subcaption{BCO}
        \label{fig:experiment:trajectory_of_No.4_NoF&NoW}
    \end{minipage}
    \caption{Visualization of demonstration and policy trajectories of Subject 4: This figure shows 50 demonstration trajectories and 30 policy trajectories for each method: (a) FABCO, (b) FABCO w/o weighting, (c) FABCO w/o FB, and (d) BCO.  Demonstrations of (a) FABCO and (b) FABCO w/o weighting are identical and were collected with visual feasibility feedback. Demonstrations of (c) FABCO w/o FB and (d) BCO are identical and were collected without visual feasibility feedback.}
    \label{fig:experiment:Trajectory_of_No.4}
\end{figure}

Fig. \ref{fig:experiment:Trajectory_of_No.4} shows the demonstration trajectory of Subject 4 and the trajectory generated by the learned policy.
The demonstrated motion clearly generated a trajectory from the initial position to the pipette insertion position regardless of the presence of feedback. 
However, the policy learned from the demonstration data collected without visual feedback failed to imitate the demonstrated motion due to error compounding.

\begin{figure}[t]
    \centering
    \includegraphics[width=0.4\hsize]{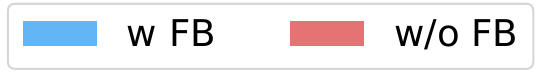}
    \includegraphics[width=0.9\hsize]{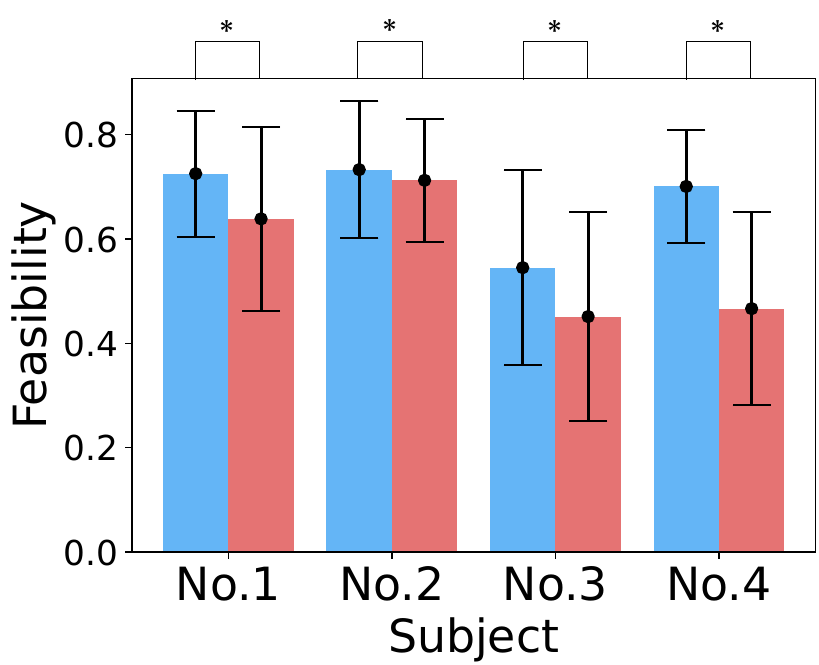}
    \caption{Comparison of feasibility with/without visual feasibility feedback for each subject's demonstration: Blue and red bars show mean and standard deviation of the feasibility across 50 demonstrations with/without visual feedback. (* means $p < 0.01$).}
    \label{fig:experiment:error}
\end{figure}

Fig. \ref{fig:experiment:error} shows the feasibility of the demonstration trajectories with/without visual feasibility feedback, indicating that using visual feasibility feedback helps demonstrate executable motion by the robot. 
Additionally, a t-test confirmed a significant difference in feasibility between the demonstrations with/without visual feedback.

\begin{figure}[t]
    \centering
    \includegraphics[width=0.4\hsize]{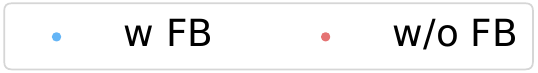}
    \includegraphics[width=0.9\hsize]{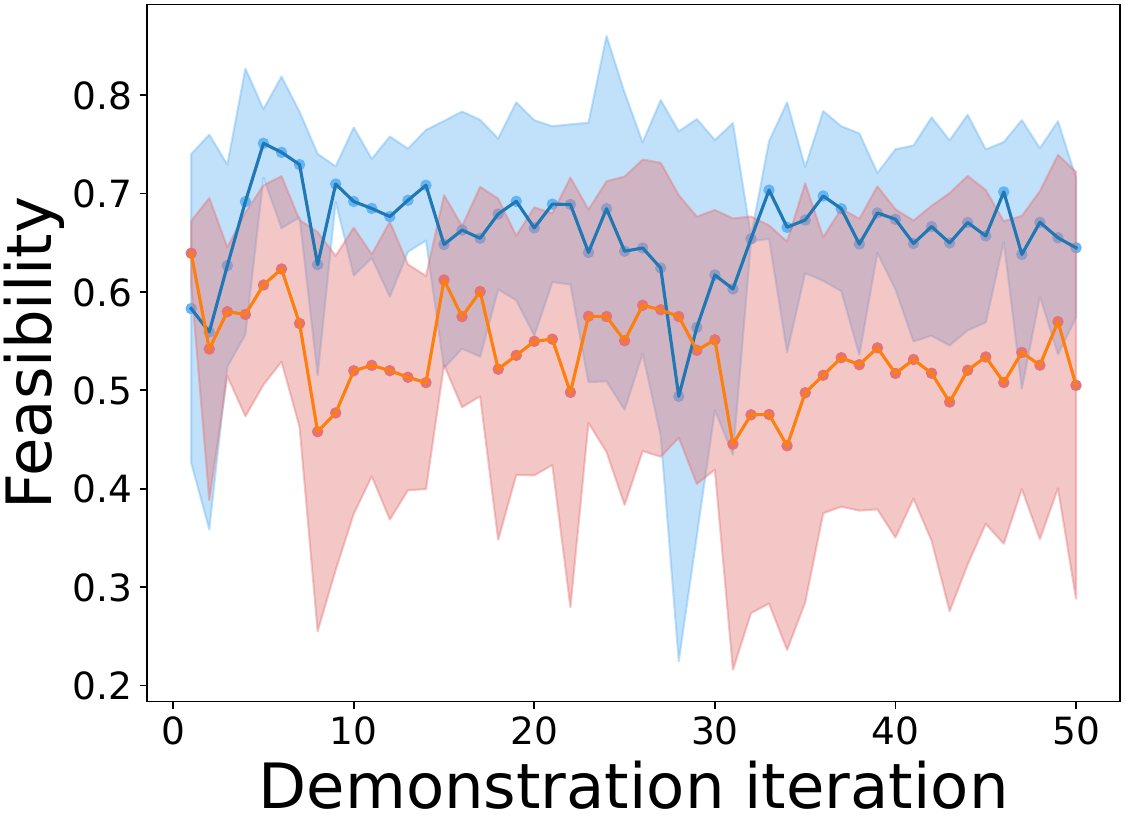}
    \caption{Transition of feasibility on demonstration: Figure shows mean and variance of feasibility at each iteration across number of demonstrations by four participants. Blue and red lines indicate transition of feasibility with/without visual feedback.}
    \label{fig:experiment:error_iteration}
\end{figure}

Fig. \ref{fig:experiment:error_iteration} illustrates the transition of feasibility over 50 demonstrations by the four participants. Demonstrations without visual feedback show low feasibility and high variance throughout; demonstrations with visual feedback start with low feasibility but improve as the participants adjust their motions based on the feedback, leading to demonstrations with higher feasibility.

The results of the NASA-TLX survey completed by the participants after the experiment are shown in Table \ref{table:experiment:NASA-TLX}. 
The NASA-TLX results indicated that while visual feedback slightly increased the demonstrators' mental demand and effort, it allowed them to perform the demonstrations with a greater sense of task accomplishment.
This suggests that although the feedback slightly increased the number of factors the demonstrators needed to consider to collect data that the robot could easily imitate, the sense of contributing valuable data to the robot's policy learning increased their sense of accomplishment.

\begin{table}[t]
    \centering
    \caption{NASA-TLX results: Higher values indicate greater workload. For each workload factor, responses are given on a scale from 0 to 9.5 in 0.5 increments. A t-test revealed a significant difference in performance score between presence/absence of visual feedback ($p < 0.01$).}

    \label{table:experiment:NASA-TLX} 
    \begin{tabular}{|c||c|c|} \hline
        NASA-TLX        & w FB            & w/o FB \\ \hline\hline
        Mental demand   & 5.50 $\pm$ 2.42 & 4.38 $\pm$ 1.85 \\ \hline
        Physical demand & 6.75 $\pm$ 0.90 & 6.88 $\pm$ 0.22 \\ \hline
        Temporal demand & 2.62 $\pm$ 1.56 & 2.88 $\pm$ 2.16 \\ \hline
        Performance     & 2.00 $\pm$ 0.87 & 5.50 $\pm$ 1.27 \\ \hline
        Effort          & 4.50 $\pm$ 2.55 & 3.88 $\pm$ 1.75 \\ \hline
        Frustration     & 2.75 $\pm$ 1.35 & 2.75 $\pm$ 1.25 \\ \hline
    \end{tabular}
\end{table}

\section{Discussion}
We identified three limitations involving the utilization of FABCO in practical applications.
The first concerns the learning of the forward and inverse dynamic models.
The dynamics model in this study is a simple neural network, which is expected to improve model performance.
By using a model that considers the robot's kinematic characteristics\cite{lutter2019deep, carron2019}, perhaps more complex movements can be achieved. 
By learning a model that considers object contact, our method may be applied to contact-rich tasks.
The second limitation is related to the feasibility calculation.
FABCO only provides feedback on feasibility, meaning that the demonstrator determines the causes of any decrease in feasibility. 
Therefore, we need to identify the factors that reduce feasibility and visualize them to the demonstrators.
The third involves the adjustment of the feasibility scale parameters.
The current method determines them heuristically.
To make FABCO more practical, they should be adjusted using demonstration and robot motion data.

\section{Conclusions}
We proposed FABCO, an imitation learning framework that encourages demonstrators to provide feasible demonstrations for robots by giving visual feedback on the feasibility of the demonstrated actions.
To validate the effectiveness of FABCO, we collected pipette insertion data from four participants, compared the feasibility with/without visual feedback as well as the performance between the policies learned using FABCO and BCO, and analyzed the mental workload using NASA-TLX. 
The results showed an improvement in feasibility with visual feedback.
Additionally, the policies learned with FABCO achieved higher task success rates than those learned with BCO. 

\bibliographystyle{IEEEtran}
\bibliography{reference}
\end{document}